\documentclass[a4paper,twoside]{article}

\usepackage{epsfig}
\usepackage{subcaption}
\usepackage{calc}
\usepackage{bm}
\usepackage{amssymb}
\usepackage{amstext}
\usepackage{textcomp}
\usepackage{amsmath}
\usepackage{amsthm}
\usepackage{multicol}
\usepackage{pslatex}
\usepackage{apalike}
\usepackage{SCITEPRESS}
\usepackage[justification=centering]{caption}

\newcommand\blfootnote[1]{%
  \begingroup
  \renewcommand\thefootnote{}\footnote{#1}%
  \addtocounter{footnote}{-1}%
  \endgroup
}

\begin{document}
\title{MAGNET: Multi-Label Text Classification using Attention-based Graph Neural Network}
\author{\authorname{Ankit Pal, Muru Selvakumar and Malaikannan Sankarasubbu}
\affiliation{Saama AI Research, Chennai, India}
\email{\{ankit.pal, selvakumar.murugan, malaikannan.sankarasubbu\}@saama.com}
}

\keywords{Multi-label text classification, Graph Neural Networks, Attention networks, Deep Learning, Natural Language Processing, Supervised Learning}

\abstract{In Multi-Label Text Classification (MLTC), one sample can belong to more than one class. It is observed that most MLTC tasks, there are dependencies or correlations among labels. Existing methods tend to ignore the relationship among labels. In this paper, a graph attention network-based model is proposed to capture the attentive dependency structure among the labels. The graph attention network uses a feature matrix and a correlation matrix to capture and explore the crucial dependencies between the labels and generate classifiers for the task. The generated classifiers are applied to sentence feature vectors obtained from the text feature extraction network(BiLSTM) to enable end-to-end training. Attention allows the system to assign different weights to neighbor nodes per label, thus allowing it to learn the dependencies among labels implicitly. The results of the proposed model are validated on five real-world MLTC datasets. The proposed model achieves similar or better performance compared to the previous state-of-the-art models.}
\onecolumn \maketitle \normalsize \setcounter{footnote}{0} \vfill

\section{\uppercase{Introduction}}
\label{sec:introduction}

\noindent   Multi-Label Text Classification (MLTC) is the task of assigning one or more labels to each input sample in the corpus. This makes it both a challenging and essential task in Natural Language Processing(NLP). We have a set of labelled training data $\{(x_{i},y_{i})\}_{i=1}^n,$ where ${x_i\; \in} \;\mathbb{R}{^D}$ are the input features with $D$ dimension for each data instances and ${y_i\; \in} \;{\{0,1}\}$ are the targets. The vector ${y_i}$ has one in the ${j}$th coordinate if the ${i}$th data point belongs to ${j}$th class. We need to learn a mapping (prediction rule) between the features and the labels, such that we can predict the class label vector ${y}$ of a new data point ${x}$ correctly.

MLTC has many real-world applications, such as text categorization \cite{DBLP:journals/ml/SchapireS00}, tag recommendation  \cite{KTV08}, information retrieval  \cite{DBLP:conf/sigir/GopalY10}, and so on. Before deep learning, the solution to the MLTC task used to focus on traditional machine learning algorithms. 

\blfootnote{ \footnotesize In Proceedings of the 12th International Conference on Agents and Artificial Intelligence (ICAART 2020) \vspace{0.7mm} \newline DOI: 10.5220/0008940304940505 \vspace{0.7mm} \newline ISBN: 978-989-758-395-7 \vspace{0.7mm} \newline Copyright \textsuperscript{\textcopyright}  2020 by SCITEPRESS – Science and Technology Publications, Lda. All rights reserved}

Different techniques have been proposed in the literature for treating multi-label classification problems. In some of them, multiple single-label classifiers are combined to emulate MLTC problems. Other techniques involve modifying single-label classifiers by changing their algorithms to allow their use in multi-label problems.

The most popular traditional method for solving MLTC is Binary Relevance (BR) \cite{Zhang2018}. BR emulates the MLTC task into multiple independent binary classification problems. However, it ignores the correlation or the dependencies among labels \cite{Luaces2012}. Binary Relevance has stimulated research for finding approaches to capture and explore the label correlations in various ways. Some methods, including Deep Neural Network (DNN) based and probabilistic based models, have been introduced to model dependencies among labels, such as Hierarchical Text Classification. \cite{DBLP:conf/icdm/SunL01}, \cite{DBLP:conf/sigir/XueXYY08}, \cite{DBLP:conf/nips/GopalYBN12} and \cite{DBLP:journals/corr/abs-1906-04898}.
Recently Graph-based Neural Networks \cite{DBLP:journals/corr/abs-1901-00596}  e.g. Graph Convolution Network \cite{DBLP:journals/corr/KipfW16}, Graph Attention Networks \cite{DBLP:conf/iclr/VelickovicCCRLB18} and Graph Embeddings \cite{DBLP:journals/corr/abs-1709-07604} have received considerable research attention. This is due to the fact that many real-world problems in complex systems, such as recommendation systems \cite{DBLP:journals/corr/abs-1806-01973}, social networks and biological networks \cite{Fout:2017:PIP:3295222.3295399} etc, can be modelled as machine learning tasks over large networks. Graph Convolutional Network (GCN) was proposed to deal with graph structures. The GCN benefits from the advantage of the  Convolutional Neural Network(CNN) architecture: it performs predictions with high accuracy, but a relatively low computational cost by utilizing fewer parameters compared to a fully connected multi-layer perceptron (MLP) model. It can also capture essential sentence features that determine node properties by analyzing relations between neighboring nodes. Despite the advantages as mentioned above, we suspect that the GCN is still missing an essential structural feature to capture better correlation or dependencies between nodes.

One possible approach to improve the GCN performance is to add adaptive attention weights depending on the feature matrix to graph convolutions.

To capture the correlation between the labels better, we propose a novel deep learning architecture based on graph attention networks. The proposed model with graph attention allows us to capture the dependency structure among labels for MLTC tasks. As a result, the correlation between labels can be automatically learned based on the feature matrix. We propose to learn inter-dependent sentence classifiers from prior label representations (e.g. word embeddings) via an attention-based function. We name the proposed method \textbf{M}ulti-label Text classification using \textbf{A}ttention based \textbf{G}raph Neural \textbf{NET}work (MAGNET). It uses a multi-head attention mechanism to extract the correlation between labels for the MLTC task.
Specifically, these are the following contributions:

\begin{itemize}
\item The drawbacks of current models for the MLTC task are analyzed.
\item  A novel end-to-end trainable deep network is proposed for MLTC. The model employs  Graph Attention Network (GAT) to find the correlation between labels.
\item  It shows that the proposed method achieves similar or better performance compared to previous State-of-the-art(SoTA) models across two MLTC metrics and five MLTC datasets.
\end{itemize}

\section{\uppercase{Related Work}}

The MLTC task can be modeled as finding an optimal label sequence ${y^*}$ that maximizes the conditional probability $p(y\mid x)$, which is calculated as follows:
\begin{equation}
    p(y\mid x) = \prod_{i=1}^n\;p(y_i\mid y_1,y_2,..,y_{i-1}, x)
\end{equation}

There are mainly three types of methods to solve the MLTC task:
\begin{itemize}
    \item Problem transformation methods
    \item Algorithm adaptation methods
    \item Neural network models
\end{itemize}
\subsection {Problem transformation methods}
Problem transformation methods transform the multi-label classification problem either into one or more single-label classification or regression problems. Most popular problem transformation method is Binary relevance (BR) (Boutell et al., 2004), BR learns a separate classifier for each label and combines the result of all classifiers into a multi-label prediction by ignoring the correlations between labels. Label Powers(LP) treats a multi-label problem as a multi-class problem by training a multi-class classifier on all unique combinations of labels in the dataset. Classifier Chains (CC) transform the multi-label text classification problem into a Bayesian conditioned chain of the binary text classification problem. However, the problem transformation method takes a lot of time and space if the dataset and labels are too large. 
\subsection {Algorithm adaptation methods}
Algorithm adaptation, on the other hand, adapts the algorithms to handle multi-label data directly, instead of transforming the data. Clare and King (2001) construct a decision tree by modifying the c4.5 algorithm \cite{Quinlan:1993:CPM:152181} and develop resampling strategies. (Elisseeff and Weston 2002) propose the Rank-SVM by amending a Support Vector Machine (SVM). (Zhang and Zhou 2007) propose a multi-label lazy learning approach (ML-KNN),  ML-KNN uses correlations of different labels by adopting the traditional K-nearest neighbor (KNN) algorithm. However, the algorithm adaptation method is limited to utilizing only the first or second order of label correlation.

\subsection {Neural network models}

In recent years, various Neural network-based models are used for MLTC task. For example, \cite{DBLP:conf/naacl/YangYDHSH16} propose hierarchical attention networks (HAN), uses the GRU gating mechanism with hierarchical attention for document classification. Zhang and Zhou (2006) propose a framework called Back-propagation for multilabel learning (BP-MLL) that learns ranking errors in neural networks via back-propagation. However, these types of neural networks don't perform well on high dimensional and large-scale data.

 Many CNN based model, RCNN \cite{DBLP:conf/aaai/LaiXLZ15}, Ensemble method of CNN and RNN by Chen et al. (2017), XML-CNN \cite{DBLP:conf/sigir/LiuCWY17}, CNN \cite{DBLP:conf/emnlp/Kim14} and TEXTCNN \cite{DBLP:conf/emnlp/Kim14}  have been proposed to solve the MLTC task. However, they neglect the correlations between labels.
 
 To utilise the  relation between the labels some Hierarchical text classification models have been proposed, Transfer learning idea proposed by \cite{DBLP:conf/icml/XiaoZW11} uses hierarchical Support Vector Machine (SVM), 
\cite{DBLP:conf/nips/GopalYBN12} and \cite{DBLP:journals/tkdd/GopalY15} uses hierarchical and graphical dependencies between class-labels, \cite{DBLP:conf/www/PengLHLBWS018} utilize the graph operation on the graph of words. However, these methods are limited as they consider only pair-wise relation due to computational constraints.

Recently, the BERT language model achieves state-of-the-art performance in many NLP tasks. \cite{Devlin2019BERTPO}

\section{\uppercase{MAGNET ARCHITECTURE}}
\subsection{Graph representation of labels}
A graph $\mathcal{G}$ consists of a feature description 
$M \in \mathbb{R}{^{n\times d}}$ and the corresponding adjacency matrix $A \in \mathbb{R}{^{n\times n}}$ where $n$ denotes the number of labels and $d$ denotes the number of dimensions.

GAT network takes the node features and adjacency matrix that represents the graph data as inputs. The adjacency matrix is constructed based on the samples. In our case, we do not have a graph dataset. Instead, we learn the adjacency matrix, hoping that the model will determine the graph, thereby learning the correlation of the labels.

Our intuition is that by modeling the correlation among labels as a weighted graph, we force the GAT network to learn such that the adjacency matrix and the attention weights together represent the correlation. We use three methods to initialize the weights of the adjacency matrix. Section 3.5 explains the initialization methods in detail.

In the context of our model, the embedding vectors of the labels act as the node features, and the adjacency matrix is a learn-able parameter.

\begin{figure*}[!h]
 \begin{center}
 \includegraphics[height=0.7 \linewidth]{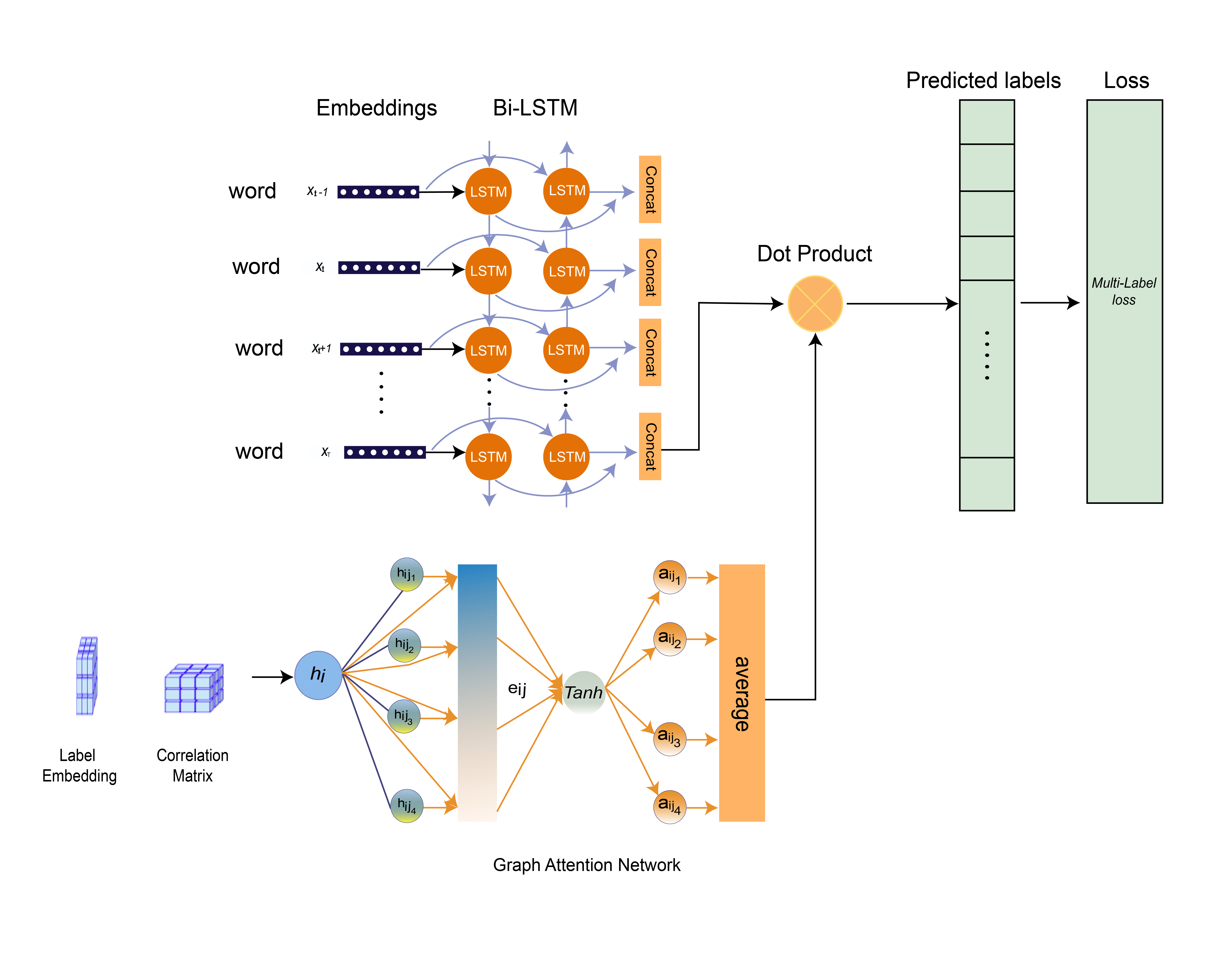}
 \caption{\textbf{Illustration of overall structure of MAGNET model with a single Graph Attention layer for multi label text classification. $\left(x^{(n)}, \mathbf{y}^{(n)}\right), n=1,2, \dots, N $ is input for BiLSTM to generate the feature vectors. $x_{(n)}$ are encoded using BERT embeddings. Input for Graph attention network is the  Adjacency matrix $A \in \mathbb{R}{^{n\times n}}$  and label vectors $M \in \mathbb{R}{^{n\times d}}$. GAT output is attended label features which is applied on the feature vectors obtained from the BiLSTM}
 \label{fig:FeatLearn1}}
 \end{center}
 \end{figure*}

\subsection{Node updating Mechanism in Graph convolution}

In Graph Convolution Network Nodes can be updated by different types of node updating mechanisms. The basic version of GCN updates each node $i$ of the ${\ell}$-th layer, $\mathbf{H_{i}^{(\bm{\ell}+1)}}$, as follows.
\begin{equation}
    \mathbf{H^{(\bm{\ell}+1)}= \sigma\left(A H^{\bm{\ell}} W^{\bm{\ell}}\right)}
\end{equation}

Where $\sigma(\cdot)$ denote an activation function, $A$ is an adjacency matrix and $W^{({\ell})}$ is the convolution weights of the ${{\ell}}$-th layer.
We represent each node of the graph structure as a label; at each layer, the label's features are aggregated by neighbors to form the label features of the next layer. In this way, features become increasingly more abstract at each consecutive layer. e.g., label 2 has three adjacent labels 1, 3 and 4. In this case, another way to write equation (2) is

\begin{multline}
\mathbf{H_{2}^{(\bm{\ell}+1)}} = \mathbf{\sigma \Bigl(H_{2}^{(\bm{\ell})} W^{(\bm{\ell})}+  H_{1}^{(\bm{\ell})} W^{(\bm{\ell})}} \\
\mathbf{+  H_{3}^{(\bm{\ell})} W^{(\bm{\ell})}+ H_{4}^{(\bm{\ell})} W^{(\bm{\ell})} \Bigr)}
\end{multline}

So, in this case, the graph convolution network sums up all labels features with the same convolution weights, and then the result is passed through one activation function to produce the updated node feature output.

\subsection{Graph Attention Networks for multi-label classification }
In GCNs, the neighborhoods of nodes combine with equal or pre-defined weights. However, the influence of neighbors can vary greatly, and the attention mechanism can identify label importance in correlation graph by considering the importance of their neighbor labels.

The node updating mechanism, equation (3), can be written as a linear combination of neighboring labels with attention coefficients.

\begin{multline}
\small 
\mathbf{H_{2}^{(\bm{\ell}+1)}} = \mathbf{ReLU \Bigl( \alpha_{22}^{(\bm{\ell})} H_{2}^{(\bm{\ell})} W^{(\bm{\ell})}+\alpha_{21}^{(\bm{\ell})} H_{1}^{(\bm{\ell})} W^{(\bm{\ell})}} \\
\mathbf{+\alpha_{23}^{(\bm{\ell})} H_{3}^{(\bm{\ell})} W^{(\bm{\ell})}+\alpha_{24}^{(\bm{\ell})} H_{4}^{(\bm{\ell})} W^{(\bm{\ell})} \Bigr)}
\end{multline}

where $\alpha_{i j}^{{\ell}}$ is an attention coefficient which measures the importance of the ${j}$th node in updating the ${i}$th node of the ${\ell}$-th hidden layer. The basic expression of the attention coefficient can be written as

\begin{equation}
\mathbf{\alpha_{ij}^{(\bm{\ell})}} = f \mathbf{\Big( H_{i}^{(\bm{\ell})} W^{(\bm{\ell})}, H_{j}^{(\bm{\ell})} W^{(\bm{\ell})}\Big)}
\end{equation}

The attention coefficient can be obtained typically by i) a similarity base, ii) concatenating features, and iii) coupling all features. We evaluate the attention coefficient by concatenating features. 

\begin{equation}
\mathbf{\alpha_{i j}}=\mathbf{ReLU\left(\left(H_{i} W\right)  \mathbin\Vert \left(H_{j} W\right)^{T}\right)}
\end{equation}
In our experiment, we are using multi-head attention \cite{DBLP:journals/corr/VaswaniSPUJGKP17}  that utilizes $K$ different heads to describe labels relationships. The operations of the layer are independently replicated $K$ times (each replication is done with different parameters), and the outputs are aggregated feature wise (typically by concatenating or adding).
\begin{small}
\begin{equation}
\mathbf{H_{i}^{(\bm{\ell}+1)}} = Tanh\left(\frac{1}{K} \sum_{k=1}^{K} \sum_{j \in N(i)} \alpha_{ij, k}^{\bm{\ell}} H_{j}^{\bm{\ell}} W^{\bm{\ell}}\right)
\end{equation}
\end{small}
Where $\alpha_{i j}$ is the attention coefficient of label $j$ to $i$. $N(i)$ represents the neighborhood of label $i$ in the graph.
We use a cascade of GAT layers. For first layer the input is label embedding matrix $M \in \mathbb{R}{^{n\times d}}$.

\begin{small}

\begin{equation}
\mathbf{H_{i}^{1}}=Tanh\left(\frac{1}{K} \sum_{k=1}^{K} \sum_{j \in N(i)} \alpha_{i j, k}^{(0)} M W^{(0)}\right)
\end{equation}
\end{small}

The output from the previous GAT layer is fed into the successive GAT layer similar to RNN but the GAT layer weights are not shared 

\begin{small}

\begin{equation}
\mathbf{H_{i}^{(\bm{\ell}+1)}}= \underbrace{Tanh\left(\frac{1}{K} \sum_{k=1}^{K} \sum_{j \in N(i)} \alpha_{i j, k}^{\bm{\ell}} H_{i}^{\bm{\ell}} W^{\bm{\ell}}\right)}_{\text{attended label features}}
\end{equation}
\end{small}
The output from the last layer is the attended label features $\mathbf{H_{gat}\in \mathbb{R}^{c\times d}}$ where $c$
denotes the number of labels and $d$ denotes the dimension of the attended label features. which is applied to the textual features from the BiLSTM.

\subsection{Feature vector generation }

We are using bidirectional LSTM (Hochreiter and Schmidhuber, 1997) to obtain the feature vectors. We use BERT for embedding the words and then feed it to BiLSTM for fine-tuning for the domain-specific task, BiLSTM reads the text sequence $x$ from both directions and computes the hidden states for each word,

\begin{equation}
\begin{aligned}
\overrightarrow{\boldsymbol{h}}_{i}=\overrightarrow{\mathrm{LSTM}}\left(\overrightarrow{\boldsymbol{h}}_{i-1}, \boldsymbol{x}_{i}\right) \\
\overleftarrow{\boldsymbol{h}}_{i}=\overleftarrow{\mathrm{LSTM}}\left(\overleftarrow{\boldsymbol{h}}_{i+1}, \boldsymbol{x}_{i}\right)
 \end{aligned}
\end{equation}

We obtain the final hidden representation of the ${i}$-th word by concatenating the hidden states from both directions, 
\begin{equation}
\boldsymbol{h}_{i}=\left[\overrightarrow{\boldsymbol{h}_{i}} ; \overleftarrow{\boldsymbol{h}}_{i}\right]
\end{equation}

\begin{equation}
    \mathbf{F = f_{rnn}(f_{BERT}(s;\theta_{BERT});\theta_{rnn})\in\mathbb{R}^{D}}
\end{equation}

Where $s$ is the sentence, $\theta_{rnn}$ is Rnn parameters, $\theta_{BERT}$ is BERT’s parameter, $D$ is hidden size of BiLSTM. Later we multiply feature vectors with attended label features to get the final prediction score as,
\begin{equation}
\hat{\boldsymbol{y}}=\boldsymbol{F} \mathbf{\odot} \boldsymbol{H_{gat}}
\end{equation}
Where $\mathbf{H_{gat}\in \mathbb{R}^{c\times d}}$ and $F$ is feature vectors obtained from BiLSTM model. Figure 1 shows the overall structure of the proposed model.

\subsection{Adjacency matrix generation }
In this section, we explain how to initialize the adjacency matrix for the GAT network. We use three different methods to initialize the weights..
\begin{itemize}

 \item \textbf{Identity Matrix}
We use the Identity matrix as the adjacency matrix. Ones on the main diagonal and zeros elsewhere, i.e., starting with zero correlation as a starting point.
 
 \item \textbf{Xavier initialization}
 We use Xavier initialization \cite{DBLP:journals/jmlr/GlorotB10} to initialize the weight of adjacency matrix.
 \begin{equation}
\pm \frac{\sqrt{6}}{\sqrt{n_{i}+n_{i+1}}}
\end{equation}

where $n_{i}$ is the number of incoming network connections.

 \item \textbf{Correlation matrix }

The correlation matrix is constructed by counting the pairwise co-occurrence of labels. The frequency vector is vector $F \in \mathbb{R}{^{n}}$ where $n$ is the number of labels and $F_{i}$ is the frequency of label $i$ in the overall training set. The co-occurrence matrix is then normalized by the frequency vector.

\begin{equation}
\textit{ A = M / F}
\end{equation}

where $M \in \mathbb{R}{^{n\times n}}$ is the co-occurrence matrix and $F \in \mathbb{R}{^{n}}$ is the frequency vector of individual labels. This is similar to how the correlation matrix built-in \cite{ML-GCN_CVPR_2019}, except we do not employ binarization.

\end{itemize}

\subsection{Loss function}

We use Cross-entropy as the loss function. If the ground truth label of a data point is $y\in \mathbb{R}^{c}$ , where ${y}_{i} = \{0, 1\}$ 

\begin{equation}
\small
\mathbf{\mathcal{L}=\sum_{c=1}^{C} y^{c} \log \left(\sigma\left(\hat{y}^{c}\right)\right)+\left(1-y^{c}\right) \log \left(1-\sigma\left(\hat{y}^{c}\right)\right)}
\end{equation}
Where $\sigma$ is sigmoid activation function
\section{\uppercase{Experiment}}
\noindent In this section, we introduce the datasets, experiment details, and baseline results. Subsequently, the authors make a comparison of the proposed methods with baselines

\subsection{Datasets}
In this section, we provide detail and use the source of the datasets in the experiment. Table 2 shows the Statistics of all datasets. \newline
\textbf{Reuters-21578} is a collection of documents collected from Reuters News Wire in 1987. The Reuters-21578 test collection, together with its earlier variants, has been such a standard benchmark for the text categorization (TC) \cite{Debole:2005:ARH:1059467.1059472}.  It contains 10,788 documents, which has 8,630 documents for training and 2,158 for testing with a total of 90 categories. \newline
\textbf{RCV1-V2} provided by Lewis et al. (2004) \cite{Lewis:2004:RNB:1005332.1005345}, consists of categorized newswire stories made available by Reuters Ltd. Each newswire story can have multiple topics assigned to it, with 103 topics in total. RCV1-V2 contains 8,04,414 documents which are divided into 6,43,531 documents for training and 1,60,883 for testing.\newline
\textbf{Arxiv Academic Paper Dataset (AAPD)} is provided by Yang et al. (2018). The dataset consists of the abstract and its corresponding subjects of 55,840 academic papers, and each paper can have multiple subjects. The target is to predict subjects of an academic paper according to the content of the abstract. The AAPD dataset then divides into 44,672 documents for training and 11,168 for testing with a total of 54 classes.\newline
\textbf{Slashdot} dataset was collected from the Slashdot website and consists of article blurbs labeled with the subject categories. Slashdot contains 19,258 samples for training and 4,814 samples for testing with a total of 291 classes.\newline
\textbf{Toxic comment dataset}, We are using toxic comment dataset from Kaggle. This dataset has large number of comments from Wikipedia talk page edits. Human raters have labeled them for toxic behavior.

\subsection{Experiment details } We implement our experiments in Tensorflow on an NVIDIA 1080Ti GPU. Our model consists of two GAT layers with multi-head attention. Table 1 shows the hyper-parameters of the model on five datasets. For label representations, we adopt 768 dim BERT trained on Wikipedia and BookCorpus. For the categories whose names contain multiple words, we obtain the label representation as to the average of embeddings for all words. For all datasets, the batch size is set to 250, and out of vocabulary(OOV) words are replaced with \textit {unk}. We use BERT embedding to encode the sentences. We use Adam optimizer to minimize the final objective function. The learning rate is initialized to 0.001 and we make use of the dropout 0.5 (Srivastava et al. 2014) to avoid overfitting and clip the gradients (Pascanu, Mikolov, and Bengio 2013) to the maximum norm of 10.

\begin{table*}
\begin{center}
\begin{tabular}{ |p{3cm}|p{2cm}|p{2cm}|p{2cm}|p{2cm}|}
 \hline
\textbf{Dataset}&\textbf{Vocab Size} & \textbf{Embed size} & \textbf{Hidden size} & \textbf{Attention heads}\\
 \hline
Reuters-21578   & 20,000   & 768 & 250 & 4\\
RCV1-V2   & 50,000 & 768    &250 & 8\\
AAPD & 30,000 &  768 &  250 & 8\\
Slashdot & 30,000    & 768  & 300 & 4 \\
Toxic Comment & 50,000  & 768  & 200 & 8\\
\hline
\end{tabular}
\caption{Main experimental hyper-parameters}
\end{center}
\end{table*}

\subsection{Performance Evaluation}

\textbf{miF1} In the micro-average method, the individual true positives, false positives, and false negatives of the system are summed up for different sets and applied to get Micro-average F-Score.

\begin{equation}
\begin{aligned}
F 1-{Score_{micro} =\frac{\sum_{j=1}^{L} 2 t p_{j}}{\sum_{j=1}^{L}\left(2 t p_{j}+f p_{j}+f n_{j}\right)}} \\
{ Precision_{micro}} = \frac{\sum_{j=1}^{L} t p_{j}}{\sum_{j=1}^{L} t p_{j}+f p_{j}} \\
{Recall_{micro}=  \frac{\sum_{j=1}^{L} t p_{j}}{\sum_{j=1}^{L} t p_{j}+f n_{j}}}
\end{aligned}
\end{equation}

\textbf{Hamming loss (HL):} Hamming-Loss is the fraction of labels that are incorrectly predicted. \cite{DBLP:conf/ipmu/Destercke14}. Therefore, hamming loss takes into account the prediction of both an incorrect label and a missing label normalized over the total number of classes and the total number of examples.

\begin{table*}
\begin{center}
\begin{tabular}{ |p{3cm}|p{2cm}|p{2cm}|p{2cm}|p{2cm}|p{2cm}| }

 \hline
 \textbf{Dataset}&\textbf{Domain} &\textbf{\#Train} & \textbf{\#Test } & \textbf{Labels}\\
 \hline
Reuters-21578   & Text    &8,630 & 2,158 & 90 \\
RCV1-V2   & Text & 6,43,531    &1,60,883 & 103 \\
AAPD & Text &  44,672 &  11,168 & 54\\
Slashdot & Text    & 19,258  & 4,814 & 291 \\
Toxic Comment & Text   & 126,856  & 31,714 & 7 \\
\hline
\end{tabular}
\caption{Statistics of the datasets}
\end{center}
\end{table*}
\begin{equation}
\textit{{HL}} = \frac{1}{|N| \cdot|L|} \sum_{i=1}^{|N|} \sum_{j=1}^{|L|} \operatorname{xor}\left(y_{i, j}, z_{i, j}\right)
\end{equation}
where $y_{i, j}$ is the target and $z_{i, j}$ is the prediction. Ideally, we would expect Hamming loss, ${HL = 0}$, which would imply no error; practically the smaller the value of $\textit{hamming loss}$, the better the performance of the learning algorithm.\newline

\subsection{Comparison of methods}

We compare the performance of 27 algorithms, including state-of-the-art models. Furthermore, we compare the latest state-of-the-art models on the rcv1-v2 dataset. Compared algorithms can be categorized into three groups, as described below:

\begin{itemize}

\item \textbf{Flat baselines}
Flat Baseline models transform the documents and extract the features using \textbf{TF-IDF} \cite{Ramos_usingtf-idf},  later use those features as input to Logistic regression \textbf{(LR)} \cite{Allison:1999:LRU:1408116}  , Support Vector Machine \textbf{(SVM)}\cite{Hearst:1998:SVM:630302.630387}  , Hierarchical Support Vector Machine\textbf{ (HSVM)} \cite{Vural:2004:HMM:1015330.1015427} , Binary Relevance \textbf{(BR)}\cite{BOUTELL20041757}, Classifier \textbf{Chains}(CC)\cite{DBLP:journals/ml/ReadPHF11}. Flat Baseline methods ignore the relation between words and dependency between labels.

\item \textbf{Sequence, Graph and N-gram based models }

These types of models first transform the text dataset into sequences of words, the graph of words or N-grams features, later apply different types of deep learning models on those features including 
\textbf{CNN} \cite{DBLP:journals/corr/Kim14f},
\textbf{CNN-RNN} \cite{7966144}, 
\textbf{RCNN} \cite{DBLP:conf/aaai/LaiXLZ15},
 \textbf{DCNN} \cite{DBLP:conf/eacl/SchwenkBCL17},
\textbf{XML-CNN} \cite{DBLP:conf/sigir/LiuCWY17}, 
\textbf{HR-DGCNN} \cite{DBLP:conf/www/PengLHLBWS018},
Hierarchical LSTM \textbf{(HLSTM)} \cite{DBLP:conf/emnlp/ChenSTLL16},
multi-label classification approach based on a conditional cyclic directed graphical model \textbf{(CDN-SVM)} \cite{DBLP:conf/ijcai/GuoG11}, 
Hierarchical Attention Network \textbf{(HAN)} \cite{DBLP:conf/naacl/YangYDHSH16} and 
Bi-directional Block Self-Attention Network \textbf{(Bi-BloSAN)} \cite{DBLP:journals/corr/abs-1804-00857} etc. for the multi-label classification task For example, Hierarchical Attention Networks for Document Classification \textbf{(HAN)} uses a GRU grating mechanism to encode the sequences and apply word and sentence level attention on those sequences for document classification. Bi-directional Block Self-Attention Network \textbf{(BI-BloSAN)} uses intra-block and inter-block self-attentions to capture both local and long-range context dependencies by splitting the sequences into several blocks.

\item  \textbf{Recent state-of-the-art models }

We compare our model with different state-of-the-art models for multi-label classification task including 
\textbf{BP-MLL$_{RAD}$} \cite{DBLP:conf/pkdd/NamKMGF14}, Input Encoding with Feature Message Passing \textbf{(FMP)} \cite{DBLP:journals/corr/abs-1904-08049}, \textbf{TEXT-CNN}\cite{DBLP:conf/emnlp/Kim14}, Hierarchical taxonomy-aware and attentional graph capsule recurrent CNNs framework \textbf{(HE-AGCRCNN)}\cite{DBLP:journals/corr/abs-1906-04898}, 
\textbf{BOW-CNN}\cite{DBLP:journals/corr/Johnson014}, \textbf{Capsule-B
networks}\cite{DBLP:journals/corr/abs-1804-00538}, Hierarchical Text Classification with Reinforced Label Assignment \textbf{(HiLAP)}\cite{DBLP:journals/corr/abs-1908-10419}, Hierarchical Text Classification with Recursively Regularized Deep Graph-CNN \textbf{(HR-DGCNN)}\cite{DBLP:conf/www/PengLHLBWS018}, 
Hierarchical Transfer Learning-based Strategy \textbf{(HTrans)}\cite{DBLP:conf/acl/BanerjeeAPT19}, 
\textbf{BERT} (Bidirectional Encoder Representations from Transformers)\cite{Devlin2019BERTPO}, 
\textbf{BERT-SGM}\cite{BERT_model}, For example \textbf{FMP + LaMP} is a variant of \textbf{LaMP} model which uses Input Encoding with Feature Message Passing (FMP). It achieves state-of-the-art accuracy across five metrics and seven datasets. \textbf{HE-AGCRCNN}  uses a hierarchical taxonomy embedding method to learn the hierarchical relations among the labels.is another recent state-of-the-art model, which has shown outstanding performance in large-scale multi-label text classification. It uses a hierarchical taxonomy embedding method to learn the hierarchical relations among the labels.
\textbf{BERT} (Bidirectional Encoder Representations from Transformers) is a recent pre-trained language model that has shown groundbreaking results in many NLP tasks. BERT uses attention mechanism \textbf{(Transformer)} to learns contextual relations between words in a text.

\end{itemize}
\section{\uppercase{Performance Analysis}}
In this section, we will compare our proposed method with baselines on the test sets. Table 4 shows the detailed Comparisons of Micro F1-score for various state-of-the-art models.

\subsection{Comparisons with State-of-the-art}

First, we compare the result of Traditional Machine learning algorithms. Among  LR, SVM, and HSVM, HSVM performs better than the other two. HSVM uses SVM at each node of the Decision tree. Later we compare the result of Hierarchical, CNN based models and graph-based deep learning models. Among Hierarchical Models HLSTM, HAN, HE-AGCRCNN, HR-DGCNN, HiLAP, and HTrans, HE-AGCRCNN performs better compared to other Hierarchical models. HAN and HLSTM methods are based on recurrent neural networks. While analyzing the performance of the recurrent model with baseline Flat models, recurrent neural networks perform worse than HSVM even though there was an ignorance of label dependency in baseline models. RNN faces the problem of vanishing gradients and exploding gradients when the sequences are too long. Graph model, HR-DGCNN, performs better than recurrent and baseline models. Comparing the CNN-based model RCNN, XML-CNN, DCNN, TEXTCNN, CNN, and CNN-RNN, TEXTCNN performs better among all of them while RCNN performs worse among them.

The sequence generator model treats the multi-label classification task as a sequence generation. When comparing the sequence generator models SGM-GE and seq2seq, SGM performs better than the seq2seq network. SGM utilizes the correlation between labels by using sequence generator model with a novel decoder structure.

Comparing the proposed MAGNET against the state-of-the-art models, MAGNET significantly improved previous state-of-the-art results, we see {\fontfamily{ptm}\selectfont\texttildelow}20$\%$ improvement in miF1  comparison to HSVM model. While comparing with the best Hierarchical text classification models, we observe {\fontfamily{ptm}\selectfont\texttildelow}11$\%$, {\fontfamily{ptm}\selectfont\texttildelow}19$\%$, {\fontfamily{ptm}\selectfont\texttildelow}5$\%$ and  {\fontfamily{ptm}\selectfont\texttildelow}8$\%$  accuracy improvement compared to HE-AGCRCNN, HAN, HiLAP, HTrans respectively. The proposed model produced a {\fontfamily{ptm}\selectfont\texttildelow}16$\%$ improvement in miF1 over the popular bi-directional block self-attention network (Bi-BloSAN). 

Comparing with CNN group models, proposed model improves the performance by {\fontfamily{ptm}\selectfont\texttildelow}12$\%$ and  {\fontfamily{ptm}\selectfont\texttildelow}6$\%$ accuracy compared with TEXTCNN and BOW-CNN method respectively. MAGNET achieves {\fontfamily{ptm}\selectfont\texttildelow}2$\%$ improvement
 over state-of-the-art BERT model.

\subsection{Evaluation on other datasets}

We also evaluate our proposed model on four different datasets rather than RCV1 to observe the performance of the model on those datasets, which vary in the number of samples and the number of labels. Table 3 shows the miF1 scores for different datasets, and we also report the Hamming loss in Table 5. Evaluation results show that proposed methods achieve the best performance in the primary evaluation metrics. We observe ~3$\%$ and 4$\%$ miF1 improvement in AAPD and Slashdot dataset, respectively, as compared to the CNN-RNN method.

\begin{table*}
\begin{center}
\begin{tabular}{ |p{2cm}|p{2cm}|p{2cm}|p{2cm}|p{2cm}| }
 \hline
 \multicolumn{5}{|c|}{F1-accuracy} \\
 \hline
 \textbf{Methods}& \textbf{Reuters-21578} &\textbf{AAPD}& \textbf{Slashdot}& \textbf{Toxic} \\
 \hline
 BR   & 0.878    &0.648&0.486 & 0.853 \\
 
 BR-support & 0.872 & 0.682& 0.516 & 0.874\\
 CC   & 0.879  & 0.654   &0.480 & 0.893\\
CNN & 0.863&  0.664 &0.512 & 0.775\\
CNN-RNN & 0.855    &0.669 &0.530 & 0.904\\
\hline
 \textbf{MAGNET}& \textbf{0.899} &\textbf{0.696} & \textbf{0.568} & \textbf{0.930}\\
  \hline

\end{tabular}
\caption{Comparisons of Micro F1-score for various models on four benchmark datasets. }
\end{center}
\end{table*}

 \subsection{Analysis and Discussion}
Here we discuss a further analysis of the model and experimental results. We report the evaluation results in terms of hamming loss and  macro-F1 score. We are using a moving average with a window size of 3 to draw the plots to make the scenarios more comfortable to read.

  \subsubsection{\normalfont {\textbf{Impact of initialization of the adjacency matrix}}}
We initialized the adjacency matrix in three different ways random, identity, and co-occurrence matrix. We hypothesized that the co-occurrence matrix would perform the best since the model is fed with richer prior information than the identity matrix, where the correlation is zero and random matrix. To our surprise, random initialization performed the best at 0.887, and identity matrix performed the worst at 0.865, whereas the co-occurrence matrix achieved the micro-F1 score of 0.878. Even though Xavier initializer performed the best, all the other random initializers performed better than co-occurrence and identity matrices. This shows that the textual information from samples contain richer information than that in the label co-occurrence matrix that we initialize the adjacency with, and both co-occurrence and identity matrix, traps the model in a local minima.
 
\begin{table}
\begin{center}
\begin{tabular}{ |p{3cm}|p{2cm}| }

 \hline
  \multicolumn{2}{|c|}{Rcv1-v2} \\
 \hline
 \textbf{Method}& \textbf{Accuracy}\\
 \hline

LR&0.692 \\
SVM&0.691\\
HSVM&0.693\\
\hline
HLSTM&0.673 \\
RCNN&0.686 \\
XML-CNN&0.695 \\
HAN&0.696 \\
Bi-BloSAN&0.72 \\
DCNN&0.732 \\
SGM+GE&0.719 \\
CAPSULE-B&0.739 \\
CDN-SVM&0.738 \\
\hline
HR-DGCNN&0.761 \\
TEXTCNN&0.766 \\
HE-AGCRCNN&0.778 \\
BP-MLL$_{RAD}$&0.780 \\
HTrans&0.805 \\
BOW-CNN&0.827 \\
HilAP&0.833 \\
\hline
BERT&0.864 \\
BERT + SGM&0.846 \\
FMP + LaMP$_{pr}$&0.877 \\
\hline
 \textbf{MAGNET} &\textbf{0.885} \\
  \hline
\end{tabular}
\caption{  Comparisons of Micro F1-score for various state-of-the-art models on Rcv1-v2 dataset.}
\end{center}
\end{table}

\begin{table*}
\begin{center}
\begin{tabular}{ |p{2cm}|p{2cm}|p{2cm}|p{2cm}|p{2cm}|p{2cm}| }
 \hline
 \multicolumn{6}{|c|}{Hamming-loss} \\
 \hline
 \textbf{Methods}& \textbf{Rcv1-v2} &\textbf{AAPD} &\textbf{Reuters-21578} & \textbf{Slashdot}& \textbf{Toxic}\\
 \hline
 BR   & 0.0093    &0.0316&   0.0032 & 0.052 & 0.034 \\
 CC   & 0.0089  & 0.0306   &0.0031 & 0.057 & 0.030 \\
CNN & 0.0084&  0.0287 & 0.0033 & 0.049 & 0.039 \\
CNN-RNN & 0.0086    &0.0282 & 0.0037 &  0.046 & 0.025 \\
\hline
 \textbf{MAGNET}& \textbf{0.0079} &\textbf{0.0252} &\textbf{0.0029} & \textbf{0.039} & \textbf{0.022}\\
  \hline

\end{tabular}
\caption{ \centering Comparisons of hamming loss for various models on four benchmark datasets.\newline the smaller the value, the better.}
\end{center}
\end{table*}

\subsubsection{Results on different types of word embeddings}
In this section, we investigate the impact of the four different word embeddings on our proposed architecture, namely the Word2Vec embeddings\cite{DBLP:conf/nips/MikolovSCCD13}, Glove embeddings \cite{DBLP:conf/emnlp/PenningtonSM14}, Random embeddings, BERT embeddings \cite{DBLP:conf/naacl/DevlinCLT19}. Figure (2) and Figure (3) shows the f1 score of all four different word embeddings on the (unseen) test dataset of Reuters-21578.

 \begin{figure}[!h]
 \begin{center} \includegraphics[width=1.0\linewidth]{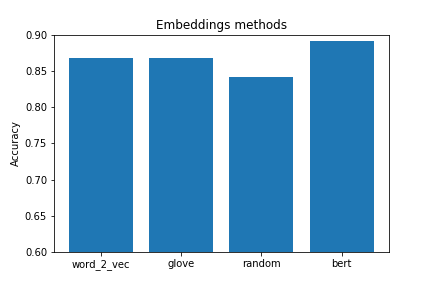}
 \caption{Different types of word embeddings performance on MAGNET x axis refer to the different types of word embeddings and y axis refer to the Accuracy ( F1-score)}
 \label{fig:FeatLearn2}
 \end{center}
 \end{figure}

Accordingly, we make the following observations:
 \begin{itemize}
      \item Glove and word2vec embeddings produce similaer results.
      
     \item Random embeddings perform worse than other embeddings. Pre-trained word embeddings have proven to be highly useful in our proposed architecture compared to the random embeddings.
     
     \item BERT embeddings outperform other embeddings in this experiment. Therefore, using BERT feature embeddings increase the accuracy and performance of our architecture.

 \end{itemize}
 Our proposed model uses BERT embeddings for encoding the sentences.

 \begin{figure}[!h]
 \begin{center}
 \includegraphics[width= 1.0\linewidth]{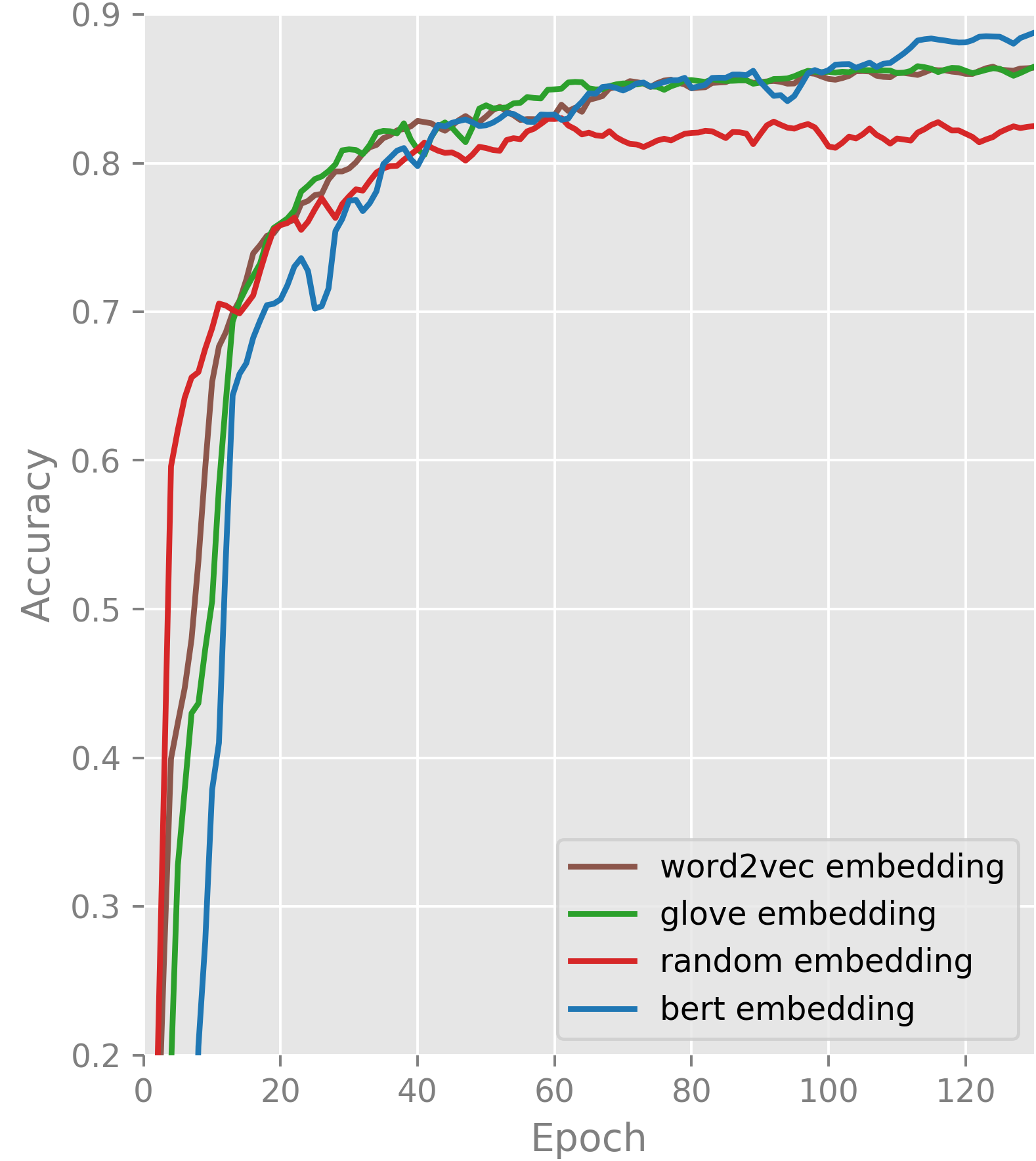}
 \caption{Performance of proposed model on different types of word embeddings. x-axis is the number of epoch and the y-axis refers to the micro-F1 score.}
 \label{fig:FeatLearn3}
 \end{center}
 \end{figure}

\subsubsection{\normalfont {\textbf{Comparison between Two different Graph neural networks}}}
In this section, we compare the performance of GAT and GCN networks. The critical difference between GAT and GCN is how the information aggregates from the neighborhood. GAT computes the hidden states of each node by attending over its neighbors, following a self-attention strategy where GCN produces the normalized sum of the node features of neighbors.

GAT improved the average miF1 score by 4$\%$ over the GCN model. It shows that the GAT model captures better label correlation compare to GCN. The attention mechanism can identify label importance in correlation graph by considering the significance of their neighbor labels.

 \begin{figure}[!h]
 \begin{center}
 \includegraphics[width=1.0\linewidth]{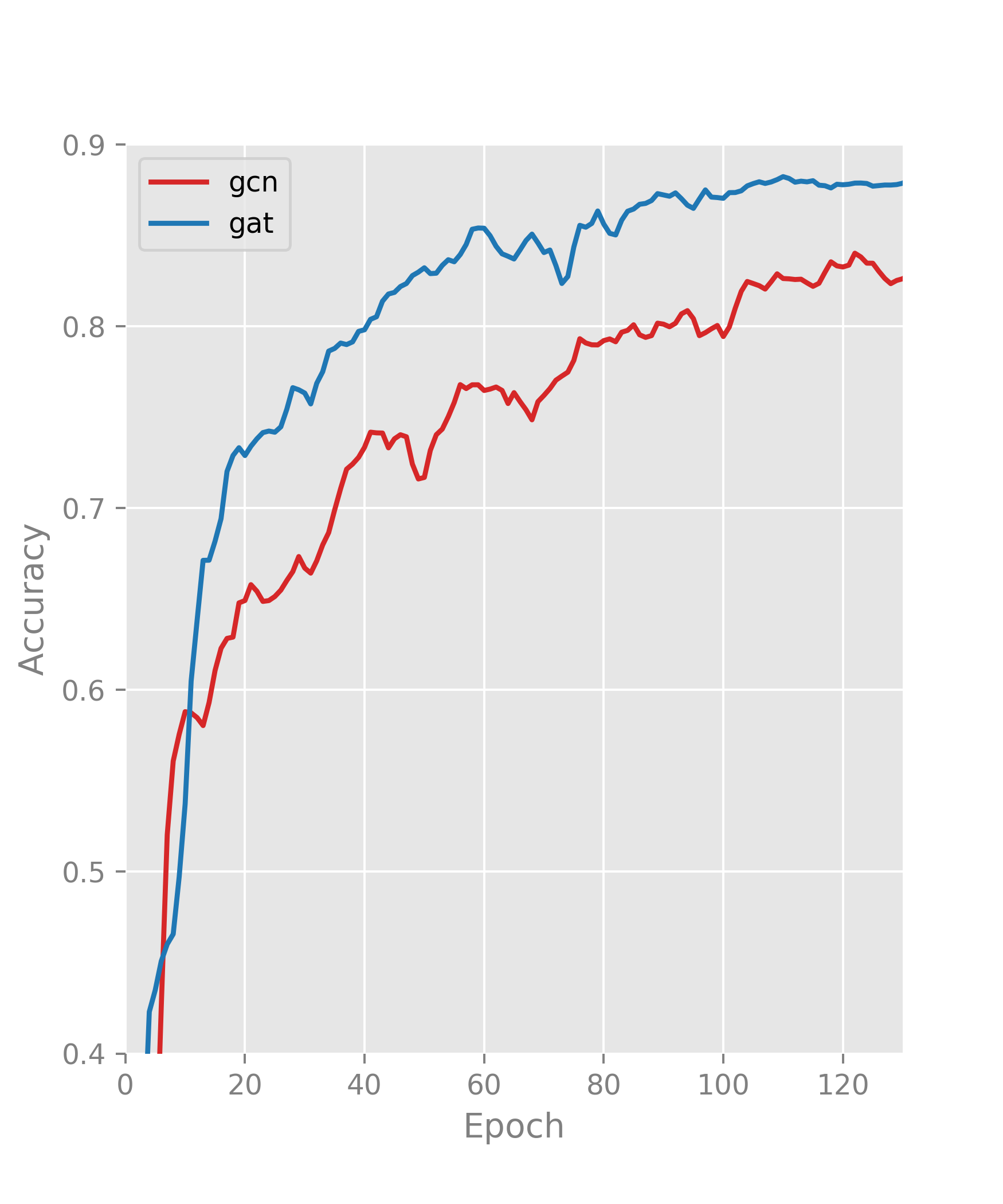}
 \caption{Performance of GAT vs GCN. x-axis is number of epochs and y-axis is micro-F1 score.}
 \label{fig:FeatLearn4}
 \end{center}
 \end{figure}
Figure (4) shows the accuracy of both neural network on Reuters-21578 dataset.
\subsection{Conclusion}

The proposed approach can improve the accuracy and efficiency of models and can work across a wide range of data types and applications. To model and capture the correlation between labels, we proposed a GAT based model for multi-label text classification.

We evaluated the proposed model on various datasets and presented the results. The combination of GAT with bi-directional LSTM shows that it has achieved consistently higher accuracy than those obtained by conventional approaches. 

Even though our proposed model performs very well, there are still some limitations. When the dataset contains a large number of labels correlation matrix will be very large, and training the model can be difficult. Our work alleviates this problem to some extent, but we still think the exploration of more effective solutions is vital in the future.

\bibliographystyle{apalike}
{\small
\bibliography{example}}

\begin{thebibliography}{}

\bibitem[Allison, 1999]{Allison:1999:LRU:1408116}
Allison, P. (1999).
\newblock {\em Logistic Regression Using Sas\textregistered: Theory and
  Application}.
\newblock SAS Publishing, first edition.

\bibitem[Banerjee et~al., 2019]{DBLP:conf/acl/BanerjeeAPT19}
Banerjee, S., Akkaya, C., Perez{-}Sorrosal, F., and Tsioutsiouliklis, K.
  (2019).
\newblock Hierarchical transfer learning for multi-label text classification.
\newblock In {\em Proceedings of the 57th Conference of the Association for
  Computational Linguistics, {ACL} 2019, Florence, Italy, July 28- August 2,
  2019, Volume 1: Long Papers}, pages 6295--6300.

\bibitem[Boutell et~al., 2004]{BOUTELL20041757}
Boutell, M.~R., Luo, J., Shen, X., and Brown, C.~M. (2004).
\newblock Learning multi-label scene classification.
\newblock {\em Pattern Recognition}, 37(9):1757 -- 1771.

\bibitem[Cai et~al., 2017]{DBLP:journals/corr/abs-1709-07604}
Cai, H., Zheng, V.~W., and Chang, K.~C. (2017).
\newblock A comprehensive survey of graph embedding: Problems, techniques and
  applications.
\newblock {\em CoRR}, abs/1709.07604.

\bibitem[{Chen} et~al., 2017]{7966144}
{Chen}, G., {Ye}, D., {Xing}, Z., {Chen}, J., and {Cambria}, E. (2017).
\newblock Ensemble application of convolutional and recurrent neural networks
  for multi-label text categorization.
\newblock In {\em 2017 International Joint Conference on Neural Networks
  (IJCNN)}, pages 2377--2383.

\bibitem[Chen et~al., 2016]{DBLP:conf/emnlp/ChenSTLL16}
Chen, H., Sun, M., Tu, C., Lin, Y., and Liu, Z. (2016).
\newblock Neural sentiment classification with user and product attention.
\newblock In {\em Proceedings of the 2016 Conference on Empirical Methods in
  Natural Language Processing, {EMNLP} 2016, Austin, Texas, USA, November 1-4,
  2016}, pages 1650--1659.

\bibitem[Chen et~al., 2019]{ML-GCN_CVPR_2019}
Chen, Z.-M., Wei, X.-S., Wang, P., and Guo, Y. (2019).
\newblock {Multi-Label Image Recognition with Graph Convolutional Networks}.
\newblock In {\em The IEEE Conference on Computer Vision and Pattern
  Recognition (CVPR)}.

\bibitem[Debole and Sebastiani, 2005]{Debole:2005:ARH:1059467.1059472}
Debole, F. and Sebastiani, F. (2005).
\newblock An analysis of the relative hardness of reuters-21578 subsets:
  Research articles.
\newblock {\em J. Am. Soc. Inf. Sci. Technol.}, 56(6):584--596.

\bibitem[Destercke, 2014]{DBLP:conf/ipmu/Destercke14}
Destercke, S. (2014).
\newblock Multilabel prediction with probability sets: The hamming loss case.
\newblock In {\em Information Processing and Management of Uncertainty in
  Knowledge-Based Systems - 15th International Conference, {IPMU} 2014,
  Montpellier, France, July 15-19, 2014, Proceedings, Part {II}}, pages
  496--505.

\bibitem[Devlin et~al., 2019a]{DBLP:conf/naacl/DevlinCLT19}
Devlin, J., Chang, M., Lee, K., and Toutanova, K. (2019a).
\newblock {BERT:} pre-training of deep bidirectional transformers for language
  understanding.
\newblock In {\em Proceedings of the 2019 Conference of the North American
  Chapter of the Association for Computational Linguistics: Human Language
  Technologies, {NAACL-HLT} 2019, Minneapolis, MN, USA, June 2-7, 2019, Volume
  1 (Long and Short Papers)}, pages 4171--4186.

\bibitem[Devlin et~al., 2019b]{Devlin2019BERTPO}
Devlin, J., Chang, M.-W., Lee, K., and Toutanova, K. (2019b).
\newblock Bert: Pre-training of deep bidirectional transformers for language
  understanding.
\newblock In {\em NAACL-HLT}.

\bibitem[Fout et~al., 2017]{Fout:2017:PIP:3295222.3295399}
Fout, A., Byrd, J., Shariat, B., and Ben-Hur, A. (2017).
\newblock Protein interface prediction using graph convolutional networks.
\newblock In {\em Proceedings of the 31st International Conference on Neural
  Information Processing Systems}, NIPS'17, pages 6533--6542, USA. Curran
  Associates Inc.

\bibitem[Glorot and Bengio, 2010]{DBLP:journals/jmlr/GlorotB10}
Glorot, X. and Bengio, Y. (2010).
\newblock Understanding the difficulty of training deep feedforward neural
  networks.
\newblock In {\em Proceedings of the Thirteenth International Conference on
  Artificial Intelligence and Statistics, {AISTATS} 2010, Chia Laguna Resort,
  Sardinia, Italy, May 13-15, 2010}, pages 249--256.

\bibitem[Gopal and Yang, 2010]{DBLP:conf/sigir/GopalY10}
Gopal, S. and Yang, Y. (2010).
\newblock Multilabel classification with meta-level features.
\newblock pages 315--322.

\bibitem[Gopal and Yang, 2015]{DBLP:journals/tkdd/GopalY15}
Gopal, S. and Yang, Y. (2015).
\newblock Hierarchical bayesian inference and recursive regularization for
  large-scale classification.
\newblock {\em {TKDD}}, 9(3):18:1--18:23.

\bibitem[Gopal et~al., 2012]{DBLP:conf/nips/GopalYBN12}
Gopal, S., Yang, Y., Bai, B., and Niculescu{-}Mizil, A. (2012).
\newblock Bayesian models for large-scale hierarchical classification.
\newblock In {\em Advances in Neural Information Processing Systems 25: 26th
  Annual Conference on Neural Information Processing Systems 2012. Proceedings
  of a meeting held December 3-6, 2012, Lake Tahoe, Nevada, United States},
  pages 2420--2428.

\bibitem[Guo and Gu, 2011]{DBLP:conf/ijcai/GuoG11}
Guo, Y. and Gu, S. (2011).
\newblock Multi-label classification using conditional dependency networks.
\newblock In {\em {IJCAI} 2011, Proceedings of the 22nd International Joint
  Conference on Artificial Intelligence, Barcelona, Catalonia, Spain, July
  16-22, 2011}, pages 1300--1305.

\bibitem[Hearst, 1998]{Hearst:1998:SVM:630302.630387}
Hearst, M.~A. (1998).
\newblock Support vector machines.
\newblock {\em IEEE Intelligent Systems}, 13(4):18--28.

\bibitem[Johnson and Zhang, 2014]{DBLP:journals/corr/Johnson014}
Johnson, R. and Zhang, T. (2014).
\newblock Effective use of word order for text categorization with
  convolutional neural networks.
\newblock {\em CoRR}, abs/1412.1058.

\bibitem[Katakis et~al., 2008]{KTV08}
Katakis, I., Tsoumakas, G., and Vlahavas, I. (2008).
\newblock Multilabel text classification for automated tag suggestion.
\newblock In {\em Proceedings of the ECML/PKDD 2008 Discovery Challenge}.

\bibitem[Kim, 2014a]{DBLP:conf/emnlp/Kim14}
Kim, Y. (2014a).
\newblock Convolutional neural networks for sentence classification.
\newblock In {\em Proceedings of the 2014 Conference on Empirical Methods in
  Natural Language Processing, {EMNLP} 2014, October 25-29, 2014, Doha, Qatar,
  {A} meeting of SIGDAT, a Special Interest Group of the {ACL}}, pages
  1746--1751.

\bibitem[Kim, 2014b]{DBLP:journals/corr/Kim14f}
Kim, Y. (2014b).
\newblock Convolutional neural networks for sentence classification.
\newblock {\em CoRR}, abs/1408.5882.

\bibitem[Kipf and Welling, 2016]{DBLP:journals/corr/KipfW16}
Kipf, T.~N. and Welling, M. (2016).
\newblock Semi-supervised classification with graph convolutional networks.
\newblock {\em CoRR}, abs/1609.02907.

\bibitem[Lai et~al., 2015]{DBLP:conf/aaai/LaiXLZ15}
Lai, S., Xu, L., Liu, K., and Zhao, J. (2015).
\newblock Recurrent convolutional neural networks for text classification.
\newblock In {\em Proceedings of the Twenty-Ninth {AAAI} Conference on
  Artificial Intelligence, January 25-30, 2015, Austin, Texas, {USA}}, pages
  2267--2273.

\bibitem[Lanchantin et~al., 2019]{DBLP:journals/corr/abs-1904-08049}
Lanchantin, J., Sekhon, A., and Qi, Y. (2019).
\newblock Neural message passing for multi-label classification.
\newblock {\em CoRR}, abs/1904.08049.

\bibitem[Lewis et~al., 2004]{Lewis:2004:RNB:1005332.1005345}
Lewis, D.~D., Yang, Y., Rose, T.~G., and Li, F. (2004).
\newblock Rcv1: A new benchmark collection for text categorization research.
\newblock {\em J. Mach. Learn. Res.}, 5:361--397.

\bibitem[Liu et~al., 2017]{DBLP:conf/sigir/LiuCWY17}
Liu, J., Chang, W., Wu, Y., and Yang, Y. (2017).
\newblock Deep learning for extreme multi-label text classification.
\newblock In {\em Proceedings of the 40th International {ACM} {SIGIR}
  Conference on Research and Development in Information Retrieval, Shinjuku,
  Tokyo, Japan, August 7-11, 2017}, pages 115--124.

\bibitem[Luaces et~al., 2012]{Luaces2012}
Luaces, O., D{\'i}ez, J., Barranquero, J., del Coz, J.~J., and Bahamonde, A.
  (2012).
\newblock Binary relevance efficacy for multilabel classification.
\newblock {\em Progress in Artificial Intelligence}, 1(4):303--313.

\bibitem[Mao et~al., 2019]{DBLP:journals/corr/abs-1908-10419}
Mao, Y., Tian, J., Han, J., and Ren, X. (2019).
\newblock Hierarchical text classification with reinforced label assignment.
\newblock {\em CoRR}, abs/1908.10419.

\bibitem[Mikolov et~al., 2013]{DBLP:conf/nips/MikolovSCCD13}
Mikolov, T., Sutskever, I., Chen, K., Corrado, G.~S., and Dean, J. (2013).
\newblock Distributed representations of words and phrases and their
  compositionality.
\newblock In {\em Advances in Neural Information Processing Systems 26: 27th
  Annual Conference on Neural Information Processing Systems 2013. Proceedings
  of a meeting held December 5-8, 2013, Lake Tahoe, Nevada, United States.},
  pages 3111--3119.

\bibitem[Nam et~al., 2014]{DBLP:conf/pkdd/NamKMGF14}
Nam, J., Kim, J., Loza~Menc'{i}a, E., Gurevych, I., and F{\"{u}}rnkranz, J.
  (2014).
\newblock Large-scale multi-label text classification - revisiting neural
  networks.
\newblock In {\em Machine Learning and Knowledge Discovery in Databases -
  European Conference, {ECML} {PKDD} 2014, Nancy, France, September 15-19,
  2014. Proceedings, Part {II}}, pages 437--452.

\bibitem[Peng et~al., 2019]{DBLP:journals/corr/abs-1906-04898}
Peng, H., Li, J., Gong, Q., Wang, S., He, L., Li, B., Wang, L., and Yu, P.~S.
  (2019).
\newblock Hierarchical taxonomy-aware and attentional graph capsule rcnns for
  large-scale multi-label text classification.
\newblock {\em CoRR}, abs/1906.04898.

\bibitem[Peng et~al., 2018]{DBLP:conf/www/PengLHLBWS018}
Peng, H., Li, J., He, Y., Liu, Y., Bao, M., Wang, L., Song, Y., and Yang, Q.
  (2018).
\newblock Large-scale hierarchical text classification with recursively
  regularized deep graph-cnn.
\newblock In {\em Proceedings of the 2018 World Wide Web Conference on World
  Wide Web, {WWW} 2018, Lyon, France, April 23-27, 2018}, pages 1063--1072.

\bibitem[Pennington et~al., 2014]{DBLP:conf/emnlp/PenningtonSM14}
Pennington, J., Socher, R., and Manning, C.~D. (2014).
\newblock Glove: Global vectors for word representation.
\newblock In {\em Proceedings of the 2014 Conference on Empirical Methods in
  Natural Language Processing, {EMNLP} 2014, October 25-29, 2014, Doha, Qatar,
  {A} meeting of SIGDAT, a Special Interest Group of the {ACL}}, pages
  1532--1543.

\bibitem[Quinlan, 1993]{Quinlan:1993:CPM:152181}
Quinlan, J.~R. (1993).
\newblock {\em C4.5: Programs for Machine Learning}.
\newblock Morgan Kaufmann Publishers Inc., San Francisco, CA, USA.

\bibitem[Ramos, ]{Ramos_usingtf-idf}
Ramos, J.
\newblock Using tf-idf to determine word relevance in document queries.

\bibitem[Read et~al., 2011]{DBLP:journals/ml/ReadPHF11}
Read, J., Pfahringer, B., Holmes, G., and Frank, E. (2011).
\newblock Classifier chains for multi-label classification.
\newblock {\em Machine Learning}, 85(3):333--359.

\bibitem[Schapire and Singer, 2000]{DBLP:journals/ml/SchapireS00}
Schapire, R.~E. and Singer, Y. (2000).
\newblock Boostexter: {A} boosting-based system for text categorization.
\newblock {\em Machine Learning}, 39(2/3):135--168.

\bibitem[Schwenk et~al., 2017]{DBLP:conf/eacl/SchwenkBCL17}
Schwenk, H., Barrault, L., Conneau, A., and LeCun, Y. (2017).
\newblock Very deep convolutional networks for text classification.
\newblock In {\em Proceedings of the 15th Conference of the European Chapter of
  the Association for Computational Linguistics, {EACL} 2017, Valencia, Spain,
  April 3-7, 2017, Volume 1: Long Papers}, pages 1107--1116.

\bibitem[Shen et~al., 2018]{DBLP:journals/corr/abs-1804-00857}
Shen, T., Zhou, T., Long, G., Jiang, J., and Zhang, C. (2018).
\newblock Bi-directional block self-attention for fast and memory-efficient
  sequence modeling.
\newblock {\em CoRR}, abs/1804.00857.

\bibitem[Sun and Lim, 2001]{DBLP:conf/icdm/SunL01}
Sun, A. and Lim, E. (2001).
\newblock Hierarchical text classification and evaluation.
\newblock In {\em Proceedings of the 2001 {IEEE} International Conference on
  Data Mining, 29 November - 2 December 2001, San Jose, California, {USA}},
  pages 521--528.

\bibitem[Vaswani et~al., 2017]{DBLP:journals/corr/VaswaniSPUJGKP17}
Vaswani, A., Shazeer, N., Parmar, N., Uszkoreit, J., Jones, L., Gomez, A.~N.,
  Kaiser, L., and Polosukhin, I. (2017).
\newblock Attention is all you need.
\newblock {\em CoRR}, abs/1706.03762.

\bibitem[Velickovic et~al., 2018]{DBLP:conf/iclr/VelickovicCCRLB18}
Velickovic, P., Cucurull, G., Casanova, A., Romero, A., Li{\`{o}}, P., and
  Bengio, Y. (2018).
\newblock Graph attention networks.
\newblock In {\em 6th International Conference on Learning Representations,
  {ICLR} 2018, Vancouver, BC, Canada, April 30 - May 3, 2018, Conference Track
  Proceedings}.

\bibitem[Vural and Dy, 2004]{Vural:2004:HMM:1015330.1015427}
Vural, V. and Dy, J.~G. (2004).
\newblock A hierarchical method for multi-class support vector machines.
\newblock In {\em Proceedings of the Twenty-first International Conference on
  Machine Learning}, ICML '04, pages 105--, New York, NY, USA. ACM.

\bibitem[Wu et~al., 2019]{DBLP:journals/corr/abs-1901-00596}
Wu, Z., Pan, S., Chen, F., Long, G., Zhang, C., and Yu, P.~S. (2019).
\newblock A comprehensive survey on graph neural networks.
\newblock {\em CoRR}, abs/1901.00596.

\bibitem[Xiao et~al., 2011]{DBLP:conf/icml/XiaoZW11}
Xiao, L., Zhou, D., and Wu, M. (2011).
\newblock Hierarchical classification via orthogonal transfer.
\newblock In {\em Proceedings of the 28th International Conference on Machine
  Learning, {ICML} 2011, Bellevue, Washington, USA, June 28 - July 2, 2011},
  pages 801--808.

\bibitem[Xue et~al., 2008]{DBLP:conf/sigir/XueXYY08}
Xue, G., Xing, D., Yang, Q., and Yu, Y. (2008).
\newblock Deep classification in large-scale text hierarchies.
\newblock In {\em Proceedings of the 31st Annual International {ACM} {SIGIR}
  Conference on Research and Development in Information Retrieval, {SIGIR}
  2008, Singapore, July 20-24, 2008}, pages 619--626.

\bibitem[Yang et~al., 2016]{DBLP:conf/naacl/YangYDHSH16}
Yang, Z., Yang, D., Dyer, C., He, X., Smola, A.~J., and Hovy, E.~H. (2016).
\newblock Hierarchical attention networks for document classification.
\newblock In {\em {NAACL} {HLT} 2016, The 2016 Conference of the North American
  Chapter of the Association for Computational Linguistics: Human Language
  Technologies, San Diego California, USA, June 12-17, 2016}, pages 1480--1489.

\bibitem[Yarullin and Serdyukov, 2019]{BERT_model}
Yarullin, R. and Serdyukov, P. (2019).
\newblock Bert for sequence-to-sequence multi-label text classification.

\bibitem[Ying et~al., 2018]{DBLP:journals/corr/abs-1806-01973}
Ying, R., He, R., Chen, K., Eksombatchai, P., Hamilton, W.~L., and Leskovec, J.
  (2018).
\newblock Graph convolutional neural networks for web-scale recommender
  systems.
\newblock {\em CoRR}, abs/1806.01973.

\bibitem[Zhang et~al., 2018]{Zhang2018}
Zhang, M.-L., Li, Y.-K., Liu, X.-Y., and Geng, X. (2018).
\newblock Binary relevance for multi-label learning: an overview.
\newblock {\em Frontiers of Computer Science}, 12(2):191--202.

\bibitem[Zhao et~al., 2018]{DBLP:journals/corr/abs-1804-00538}
Zhao, W., Ye, J., Yang, M., Lei, Z., Zhang, S., and Zhao, Z. (2018).
\newblock Investigating capsule networks with dynamic routing for text
  classification.
\newblock {\em CoRR}, abs/1804.00538.

\end{thebibliography}

\end{document}